# Five Dimensions of Reasoning in the Wild

**Don Perlis**
University of Maryland, College Park MD
perlis@cs.umd.edu

**Abstract**

Reasoning does not work well when done in isolation from its significance, both to the needs and interests of an agent and with respect to the wider world. Moreover, those issues may best be handled with a new sort of data structure that goes beyond the knowledge base and incorporates aspects of perceptual knowledge and even more, in which a kind of anticipatory action may be key.

## Out of the Ivory Tower

Reasoning is one of the oldest topics in artificial intelligence (AI). And it has made lots of progress, in the form of commonsense reasoning (CSR), planning, automated theorem-proving, and more. But I suspect it has hit a barrier that must be surmounted if we are to approach anything like human-level inference. Here I give evidence for such a barrier, and ideas about dealing with it, loosely based on evidence from human behavior.

In rough synopsis, reasoning does not work well when done in isolation from its broader significance, both for the needs and interests of an agent and for the wider world. Moreover, those issues may best be handled with a new sort of data structure that goes beyond the knowledge base (KB) and incorporates aspects of perceptual knowledge and even more, in which a kind of anticipatory action many be key.

I suspect this has ties with recent calls to "put the Science" back in AI (Levesque 2013, Langley 2012). For what I am arguing, in some sense, is that reasoning should be regarded as "in the wild" as events unfold rather than confined to management of an isolated KB; and that this speaks to an agent interacting with the world, rather than a puzzle in abstract inference (yet I will also argue that even "pure" reasoning as in mathematics hugely benefits from many connections with the world). And finally, we then will end up studying the nature of world-embedded cognitive agents, humans included. But this is very broad-brushed and general, whereas my main point is a technical suggestion about reasoning informed by meaning, especially meaning concerning experience and action.

---

One quick example at the outset: The Wason Selection Task (Wason 1968) shows that human inference is strongly aided when the details of the task at hand have real meaning that the subjects can relate to in terms of things that matter to them, helping keep attention on what is relevant; and this holds even when the task in the abstract is a matter of so-called pure logic. While this could be seen as a defect in human reasoning, something computers would never trip up on, I think it points in the opposite direction: inference without broader meaning is not worth much, and not worth being good at.

I will illustrate my main points with a series of examples based on the activities of proving, planning, and understanding.

## Proving

Mathematicians not only need to stop now and then for lunch or sleep. Their respite from plowing straight ahead with lemma upon lemma goes far beyond that. In fact, I would venture to claim that mathematical reasoning is much more a matter of assessing where one is, what one has done, and where one should go next, than it is single-minded carrying out of steps in proofs. Of course, there is checking of ideas, seeing if a method pans out. But even that – except in somewhat rare cases of brute computation, say of an integral – is largely again a matter of looking to see what step to take next, whether things look as expected, and so on. And equally important: whether there is evidence of sufficient progress to justify further effort in the same direction, or perhaps it is time to try something else.

Thus I argue that mathematical prowess tends to involve large amounts of broad knowledge of connections, and not so much sequential depth. While the end-product may be long, it arises from very short pieces built from fairly direct connections, and stitched together again by similarly direct (albeit sometimes not widely known) connections.

So there is a (good) kind of shallow-but-broad character to proving theorems, and any apparent depth arises out of that. (Whether a future cousin of Deep Blue will show that the pure plowing-ahead kind of proof activity can outdo mathematicians remains to be seen; I am guessing not.)

Not only that. Actual calculational steps themselves are highly symbolic in nature – and by that I do not mean devoid of meaning. On the contrary, I refer to the actual physical moving of symbols on a blackboard – or in simulation in the mind's eye (Landy&Goldstone 2009, Shepard&Metzler 1971); it is very time-and-space oriented. This applies both to the symbols themselves, and to what they stand for; the use-mention distinction is fundamental in mathematical reasoning. A manifold is not just the letter M but a spreading multi-dimensional surface that the mathematician's "eye" can peer about in, while assessing what to consider next about M. In addition, symbols get huge re-use, so M may refer to multiple things, and assessment has to work out such symbol-clashes as well. (Miller 1993) has made an encouraging start on this latter issue, but in an NLP and CSR setting; see also (Gurney et al 1997).

Finally, mathematicians can and do – frequently – question their axioms, and alter them at will. Some of the most important advances come from seeing that a different definition or construction provides deep insights that were obscured by traditional notions.

Yet none of this has even a remote presence in automated proof-systems or in CSR. Our current artifacts are not just KB managers but even worse: clueless ones that always push straight ahead, never stopping to take stock of where they have been and where they are going.

## Planning

Suppose 500 sheets of paper are neatly stacked together (in what is known as a *ream*). If the task is to load them into a printer, this can be done one sheet at a time (terrible); or (possibly) the entire ream at once, or some portion thereof. To even address this problem requires multiply representing the stack as one (a ream) and many (sheets). Moreover, specifying the problem – if it is not to be a highly distasteful exercise in frustration – will not involve stating all 499 "on" relationships between successive sheets. Finally, there might not be 500 sheets, but rather an unspecified stack of between (say) 400 and 600 sheets. This sort of problem has motivated some (Goldman 2009, Srivastava et al 2015) to address the need for planning systems that can consider loops (do-until) and other constructs that are not part of standard planners.

A sort of spatial grasp of the problem seems essential here, if a human is not to do most of the work in preparing the problem ahead of time. An agent on its own somehow must "see" that there is a stack of many items, and then assess the many possibilities for transporting them without considering each one in turn.

Here is a second example: 100 coffee cups are sitting in scattered locations on a table. They are to be moved to one end of the table. A cup can be transported safely by attaching a string to its handle and pulling gently. The string can be attached to as many as (say) 20 cups at a time at a very slight increase in cost, and pulled with no increase in cost or risk. It is clear that the best plan is to transport twenty cups at a time, over and over. But how is a planner to figure this out? Traditional planners will figure it out only by considering all possible scenarios, including all possible choices of (ordered!) subsets of up to twenty cups! This will overwhelm even the fastest planners today and in the foreseeable future.

Yet a system endowed with the ability to "think" in some suitable sense, will realize that (i) twenty cups at a time is optimal, and (ii) it does not matter a lot which twenty except that they should be reasonably close together to facilitate the string attachment. All this, I contend, requires a rich sense of how reality works, and an ability to make assessments based on that. Aspects of this can be quite naturally handled by an appropriate KB and associated inference rules (again, see Miller 1993). But noting where the cups are, and that one should avoid a bumpy area while pulling them, have to be one-off items that no amount of prior experience will have prepared one for. This takes, among other things, reasoning about sets (McCarthy 1995, Perlis 1988).

## Understanding

Patrick Winston (Winston 2011) gives an example that will be instructive here; I paraphrase and elaborate for present purposes: *A new table-saw has a red label with the warning "Don't wear gloves when using this equipment". The owner puzzles over this, wondering whether the "don't" is a typo, or if it is a macabre joke, before envisioning the activity of sawing a piece of wood with gloves on: as the gloves are imagined to approach the spinning blade, the finger of one glove is seen to catch on a sawtooth and be pulled into the blade, along with the hand trapped inside it.*

The lesson Winston takes from this is that we are storytellers, and that much of our intelligence resides in that capacity. I do not disagree, but I also would emphasize another lesson: that we are imaginers, we envision circumstances in time and space (Fauconnier 1985, Johnson-Laird 1986), and use that to inform our undertanding of situations. In effect, we run mental simulations to see what might happen.

It is very unlikely – and Winston makes this case – that we have a supply of general axioms suitable for reaching the conclusion that one should not wear gloves when using a table-saw. Gloves after all protect ones hands from cold, wet, cuts, bruises and other dangers; and a table-saw is dangerous. Yet when we *envision* the situation, we can see details that we did not know before. We now seem to un-derstand the red warning label, or at least have a plausible explanation of it.

But perhaps we did already know this, at least implicitly? We know a sawtooth has a sharp point that can catch and cut flesh, wood, most materials – so why not also a glove? Yet a glove can be very protective around sharp items such as knives. But the saw spins, and so an item caught in a tooth could be pulled along, very fast. But did one know that ahead of time? And won't the tooth simply rip through most materials (like skin) rather than pull on them? Yet gloves often have resilience, fibrous elements that tend not to rip fully and that can be grabbed and tugged. On the other hand, so does wood. There seems no way to figure this out from routine well-known "facts".

In fact, there does not seem to be a consensus among woodworkers whether one should or should not wear gloves when using a table-saw. But one can rely on recalled images of gloves and similar materials getting caught in various ways, and hands not so quickly extracted, to come to an understanding as to why it *might* be dangerous to do so – i.e., what the warning sign could be getting at. But to do this, one has to inspect those images, they are not already parsed into thousands (millions?) of fine details in a KB ready for logic-processing. And if they were, it is highly likely that the result would be inconsistent (Perlis 1997).

So now we have seen various initial aspects of our larger point, that is still emerging: reasoning is made far more effective if it is not restricted purely to syntactic manipulations of a KB, but can also avail itself of (experience-based) simulations, which it then can assess. Indeed, reasoning can initially assess whether it is worth doing simulating (imagining) in the first place, and then what weight to give it, or whether to rerun it with some changes, and so on. And this brings us to the next section.

## The 5[th] Dimension

So, where are we? It appears that reasoning about the everyday world necessarily involves a substantial amount of information about the 4-dimensional spatial and temporal structure of that world: what is near what, above what, and so on, and how things behave over time. And it gets very complicated, apparently more than can be accommodated fully in any sort of KB that we can construct. (CYC would be an apparent counterexample, but this author maintains that even CYC (see Lenat 1995) is not up to the Winston task in a general way, for reasons given above.) On the other hand, as the saying goes, a picture is worth a thousand words. Alas, computer vision is not in such a good state either. The two clearly can help each other out; and this is a topic of ongoing research.

But there is another point, also present in the three case studies above (proving, planning, understanding): an agent needs to know not only what is what, there in front of it, so to speak, but also what *could* be what if changes were made. This is key to creating proofs, making plans, understanding risks, and so on. And that – seeing into possible futures – is one of the most effortless things we do, all day long. We are not speaking of magical predictions here, but rather of simple awareness of how things are likely to go under fairly ordinary circumstances. And so I want to add one more dimension, and suggest a set of hypotheses.

1. Our brains are geared not so much for perception *per se*, but rather for anticipation or imagination of alternatives; what could be called *envisioning*. (Note that many people do very well indeed, with quite poor vision, hearing, and other deficits.)
2. This envisioning involves special data structures with some of the 4-dimensional qualities of the actual world.
3. Our brains run very fast simulations with those structures, based in part on high-level (KB) information and in part on perceptual data, to anticipate what might happen under various conditions.
4. We specify those conditions, and even specify additional details (traditionally regarded as belonging to different "levels"): pure sensory data (red spots), shapes, items not perceived, areas of high attention, interpretations and speculations (is that blood?), etc. And all this is part of these same structures, encoded in what might be called an *imaginative-reasoning markup language* (IRML).

## IRML Structures

These additional notations are a critical 5[th] element in our reasoning, and – I claim – in any real-world agent that is broadly effective at a human level. We "see" not only what is there in front of us but also (in our reasoning) whatever we want to see, groupings that interest us, elements that are not even there but that we wonder about, and so on. We switch our thoughts and speculations at will (Perlis 1987).

Such hypothesized IRML structures could be a bit like drawings with many added elements, e.g., some items circled, arrows and notes, and so on, not unlike a cartoon. The vision community sometimes refers to a cartoon-level or mid-level structure, as a halfway step in processing image data from pixel (low-level) to scene understanding (KB, or high-level). But I am suggesting that the real understanding comes at the ongoing active IRML level, not at the KB level.

Consider the word "house" that may appear within a KB item. It has no meaning in itself, and could just as well be replaced with any generic symbol, e.g., 42akb3#0. It is only when an agent imaginatively relates such KB items to, for instance, walls, rooms, distances, space one can move about in, that a kind of *house* understanding arises. What makes a KB different from any other collection of sentences? Surely how it is *used*. And the use that I am suggesting is one of the KB being used in augmentation of an internal dynamic simulation which is populated with very many scanning processes that constantly calculate what is going on.

One understands a scene not simply when one has the abstract knowledge that there are three people playing cards; that is useful and an aspect of the scene, but for many purposes far too thin. Are they seated? Men? Women? Children? Smiling? Close enough to touch each other? How many cards are on the table? Are the cards touching each other? Does the table have straight-edge sides? Is there some discoloration halfway along the nearest side? And so on, and on. There is no seeming end to the list of details that are available. Of course, computer vision researchers would be thrilled to be able to provide just "three people playing cards"; but that merely skims the surface.

Consider simply a rectangular table. What happens when one "sees" it? What does it mean to now know that, for instance, the closest edge is straight? It cannot be simply that an "image" is stored somewhere – the image data has to be suitably processed for the straightness to be determined. And indeed there are algorithms – and neural assemblies – that can do this.

So, what I propose is that instead of having this information simply passed up to a KB, the algorithms keep at it, reiterating the processing as long as attention is on the table. Let's call these processes "fast crawlers" – they "crawl" over the image again and again, and there are many of them, akin to the "quick and inflexible" modules of (Pollock 1995). They take measurements in time, dynamically triangulating like busy little surveyors. As soon as attention moves away, they stop and we are left with a mostly abstract and thin KB version that we had been looking at a straight edge. Yet the KB is hardly inactive in all this. It does a perpetual tango with – indeed I think is best thought of as part of – the IRML structures, informing and guiding what should be imagined next.

But doesn't all this speak to more evidence of poor design in the human case? Why not keep a veridical image in memory, that can be inspected again later on? Indeed that can be done – and we do it with external aids such as photographs. But any agent will still have to rerun the fast crawlers all over again. There is no other way to "see", and with it to figure out plausible outcomes. Envisioning cannot be retained except by continuing to envision.

How might this work? One thought is that an agent can have stored "videos" of past experiences, and then as needed modify and stich them together to suit present purposes. One has seen spinning sawblades, one has seen gloves, and one creates a little internal movie that runs according to learned (and possibly some inborn) rules of how the world works (physical simulation). The result will have gaps and confusions, but may be enough to provide a rich set of possibilities that can then be further explored.

Here is another example, modified from (Levesque 2013): Can a crocodile jump over a hedge? This would be very hard for a KB to provide any sort of answer to, unless specially doctored for that purpose. Levesque argues that a great deal of world knowledge is needed to answer general questions, and I would not disagree. But I think the case is even harder than that, and here I have presented the view that very often what is needed is the ability to run virtual reality shows internally: one "sees" the crocodile scuttle toward the hedge, and either it does or does not manage to jump over, based on one's current grasp of how things work – or indeed it may fail to give an unambiguous answer. But this grasp is not entirely given in the form of explicit propositional knowledge; rather much of it is implicitly embedded in the agent's fund of experience in the form of stored videos, and processed by virtual fast crawlers. Indeed, it will be a mix of explicit reasoning and visualized reasoning, which is what the IRML idea is aimed at – though as yet still very underfleshed!

Applying these ideas to, for instance, the coffee-cup planning problem above, might allow an agent to visualize the attachment of string to many cups, and this could then facilitate such things as realizing that it is best to start with the cups easiest to reach – and to *see* which cups those are. This assumes experience with attaching strings, and with cups sliding on a table, as well as general physical "unfolding" of events in time and space. So there is a lot of work involved, but well worth it in my view.

## Related work

Of course, much work has been done on hypothetical reasoning, and even in connection with planning and perception; see for instance (Gelfond & Lifschitz 1993, Lifschitz 1999, and Baral et al 1997). However, as far as I can tell, this does not address the issues in the Winston problem, nor does it appear to lend itself to simulational (real-time dynamic) reasoning.

There appears to be very limited work on image-processing related to reasoning in the fashion I am describing; see (Glasgow 1998, Chang et al 2014, Mohan et al 2012).

NLP is an area where some potential overlap with these ideas is starting to show up, partly borrowed from linguistics; indeed, there is an exciting mix of these and neuroscience. See for instance (Feldman 1985, O'Regan & Noe 2001, Bergen & Chang 2005, and Lau et al 2008).

In the planning area, in addition to work already mentioned, there are (Riddle et al 2015, Traverso et al 2015, and the much older Nirkhe 1994), all of which consider the need for more realistic representations and methods for real-world planning.

# Conclusion

I suspect that AI – CSR and planning and vision of course, but also many other areas such as NLP and even machine learning – will not approach human-level intelligence until we take far more seriously the real-world dynamic connection of agent and environment. On the other hand, I think much of the work is clear enough for us to begin to tackle. A suitable form of dynamic movement-based envisioning may be part of the solution.

The new AI watchwords could be – instead of the traditional generate and test – *envision and assess*.

# Acknowledgements

Thanks to Mike Cox and Vikas Shivashankar for enlightening discussions. This material is based in part upon work supported by ONR Grant # N00014-12-1-0430 and ARO Grant # W911NF-12-1-0471.